%% file: main.tex
\documentclass[sigconf]{acmart}
\usepackage{xcolor}    
\usepackage{bbm}
\usepackage{algorithm}
\usepackage[noend]{algpseudocode}
\usepackage{booktabs}
\usepackage{subfig}
\usepackage{xspace} 
\usepackage{adjustbox}

\usepackage{enumitem}
\setlist{leftmargin=4.5mm}

\newcommand{\cora}{\texttt{Cora}\xspace}
\newcommand{\citeseer}{\texttt{Citeseer}\xspace}
\newcommand{\pubmed}{\texttt{Pubmed}\xspace}
\newcommand{\dblp}{\texttt{DBLP}\xspace}

\newcommand{\computer}{\texttt{A-Computers}\xspace}
\newcommand{\photo}{\texttt{A-Photo}\xspace}
\newcommand{\wiki}{\texttt{Wiki-CS}\xspace}
\newcommand{\cocs}{\texttt{Co-CS}\xspace}

\newcommand{\gra}{\texttt{GRACE}\xspace}
\newcommand{\graP}{\texttt{GRACE+}\xspace}
\newcommand{\graPG}{\texttt{GRACE+(G)}\xspace}
\newcommand{\graPF}{\texttt{GRACE+(F)}\xspace}
\newcommand{\graPP}{\texttt{GRACE+(P)}\xspace}
\newcommand{\graPN}{\texttt{GRACE+(N)}\xspace}

\newcommand{\gca}{\texttt{GCA}\xspace}
\newcommand{\gcaP}{\texttt{GCA+}\xspace}

\newcommand{\gmlp}{\texttt{Graph-MLP}\xspace}
\newcommand{\gmlpP}{\texttt{Graph-MLP+}\xspace}
\newcommand{\gmlpPG}{\texttt{Graph-MLP+(G)}\xspace}
\newcommand{\gmlpPF}{\texttt{Graph-MLP+(F)}\xspace}
\newcommand{\gmlpPP}{\texttt{Graph-MLP+(P)}\xspace}
\newcommand{\gmlpPN}{\texttt{Graph-MLP+(N)}\xspace}

\AtBeginDocument{%
  \providecommand\BibTeX{{%
    \normalfont B\kern-0.5em{\scshape i\kern-0.25em b}\kern-0.8em\TeX}}}

\setcopyright{acmcopyright}
\copyrightyear{2018}
\acmYear{2018}
\acmDOI{XXXXXXX.XXXXXXX}

\acmConference[Conference acronym 'XX]{Make sure to enter the correct
  conference title from your rights confirmation emai}{June 03--05,
  2018}{Woodstock, NY}
\acmPrice{15.00}
\acmISBN{978-1-4503-XXXX-X/18/06}

\newcommand{\ym}[1]{\textcolor{blue}{#1}}

\begin{document}

\title{Enhancing Graph Contrastive Learning with Node Similarity}


 \author{Hongliang Chi}
 \email{hc443@njit.edu}
 \affiliation{%
  \institution{New Jersey Institute of Technology}
   \country{Newark, New Jersey, USA}
 }
\author{Yao Ma}
 \email{yao.ma@njit.edu}
 \affiliation{%
   \institution{New Jersey Institute of Technology}
   \country{Newark, New Jersey, USA}
 }

\renewcommand{\shortauthors}{xxxx, et al.}

\begin{abstract}
Graph Neural Networks (GNNs) have achieved great success in learning graph representations and thus facilitating various graph-related tasks. However, most GNN methods adopt a supervised learning setting, which is not always feasible in real-world applications due to the difficulty to obtain labeled data. Hence, graph self-supervised learning has been attracting increasing attention. Graph contrastive learning (GCL) is a representative framework for self-supervised learning. In general, GCL learns node representations by contrasting semantically similar nodes (positive samples) and dissimilar nodes (negative samples) with anchor nodes. Without access to labels, positive samples are typically generated by data augmentation, and negative samples are uniformly sampled from the entire graph, which leads to a sub-optimal objective. Specifically, data augmentation naturally limits the number of positive samples that involve in the process (typically only one positive sample is adopted). On the other hand, the random sampling process would inevitably select false-negative samples (samples sharing the same semantics with the anchor). These issues limit the learning capability of GCL. In this work, we propose an enhanced objective that addresses the aforementioned issues. We first introduce an unachievable ideal objective that contains all positive samples and no false-negative samples. This ideal objective is then transformed into a probabilistic form based on the distributions for sampling positive and negative samples. We then model these distributions with node similarity and derive the enhanced objective. Comprehensive experiments on various datasets demonstrate the effectiveness of the proposed enhanced objective under different settings.

\end{abstract}




\maketitle
\input{sections/introduction}
\input{sections/methodology}

\input{sections/experiment}
\input{sections/related_work}
\input{sections/conclusion}

\clearpage
\bibliographystyle{ACM-Reference-Format}
\balance
\bibliography{reference}


\end{document}

%% file: sections/introduction.tex
\section{Introduction}\label{sec:introduction}

Graphs are regarded as a type of essential data structure to represent many real-world data, such as social networks~\cite{fan2019deep, goldenberg2021social}, transportation networks~\cite{shahsavari2015short}, and chemical molecules~\cite{kearnes2016molecular, wen2021bondnet}. Many real-world applications based on these data can be naturally treated as computational tasks on graphs. To facilitate these graph-related tasks, it is essential to learn high-quality vector representations for graphs and their components. Graph neural networks (GNNs)~\cite{kipf2016semi, velivckovic2017graph, wu2019simplifying}, which generalize deep neural networks to graphs, have demonstrated their great power in graph representation learning, thus facilitating many graph-related tasks from various fields including recommendations~\cite{fan2019graph, xu2020graphsail}, natural language processing~\cite{yao2019graph, zhang2020every}, drug discovery~\cite{li2021effective, rathi2019practical}, and computer vision~\cite{garcia2017few, qasim2019rethinking}. 

Most GNN models are trained in a supervised setting, which receives guidance from labeled data. However, in real-world applications, labeled data are often difficult to obtain while unlabeled data are abundantly available~\cite{jin2020self}. Hence, to promote GNNs' adoption in broader real-world applications, it is of great significance to develop graph representation learning techniques that do not require labels. More recently, contrastive learning techniques~\cite{he2020momentum, chen2020simple, grill2020bootstrap}, which are able to effectively leverage the unlabeled data, have been introduced for learning node representations with no labels available~\cite{wu2018unsupervised}.  

Graph contrastive learning (GCL) aims to map nodes into an embedding space where nodes with similar semantic meanings are embedded closely together while those with different semantic meanings are pushed far apart. More specifically, to achieve this goal, each node in the graph is treated as an anchor node. Then, nodes with similar semantics to this anchor are identified as positive samples while those with different semantic meanings are regarded as negative samples. A commonly adopted objective for graph contrastive learning is based on InfoNCE~\cite{van2018representation,qiu2020gcc, zhu2020deep, zhu2021graph} as follows. 
{\small
\begin{align}
\mathcal{L}({v}) = -\log \frac{e^{f(v)^{\top} f\left(v^{{\prime}}\right)/\tau}}{ e^{f(v)^{\top} f\left(v^{{\prime}}\right)/\tau} + \sum_{v_s \in \mathcal{N}(v)} e^{f(v)^{\top} f\left(v_{s}\right)/\tau} },
\label{eqn:contrastive_loss} 
\end{align}
}
where $v$ is an anchor node, $v'$ is the positive sample, $\mathcal{N}(v)$ denotes the set of negative samples, and $f(\cdot)$ maps a input node to a low-dimensional representation. As the real semantics of nodes are not accessible, the single positive sample $v'$ is often generated by data augmentation that perturbs the original graph. 
On the other hand, the set of negative samples $\mathcal{N}(v)$ are often uniformly sampled from the graph~\cite{hafidi2022negative}. More details on the graph contrastive learning objective are discussed in Section~\ref{sec:preliminary}. The final objective for graph contrastive learning is typically a summation of Eq~\eqref{eqn:contrastive_loss} over all nodes. By minimizing this objective, we pull $v'$ close to $v$ while pushing any nodes in $\mathcal{N}(v)$ far apart from $v$.  

However, such an objective is not optimal for learning high-quality representations due to two major shortcomings: 1) The set of negative samples $\mathcal{N}(v)$ inevitably contains nodes with similar semantic meanings as the anchor, which are referred as the false-negative samples. In this case, minimizing the contrastive objective in Eq~\eqref{eqn:contrastive_loss} blindly pushes away the representations of false-negative samples, which impairs the quality of learned representations. Hence, removing false-negative samples has the potential to improve the performance of contrastive learning, which is also demonstrated in~\cite{chuang2020debiased}; 2) Furthermore, the contrastive objective $\mathcal{L}(v)$ only contains a single positive sample generated from data augmentation, which limits its capability to pull similar nodes together. Preferably, including more positive samples in the numerator benefits the contrastive learning process, which is verified in~\cite{khosla2020supervised}. 

An ideal contrastive objective excludes false-negative samples from the negative sample set in the denominator but contains all positive samples in the numerator (see details in Section~\ref{sec:ideal}). However, without access to the true labels, such an ideal objective is not achievable in practice. In this work, we propose an enhanced objective that approximates the ideal objective. In particular, we first transfer the ideal objective into a probabilistic form by modeling the anchor-aware distributions for sampling positive and negative samples. Intuitively, nodes with higher semantic similarity to the anchor node should have a higher probability to be selected as positive samples. Hence, we estimate these anchor-aware distributions by theoretically relating them with node similarity. Measuring node similarity is challenging since it involves both graph structure and node features, which interact with each other in a complicated way. In this work, we propose a novel strategy to model the pairwise node similarity by effectively utilizing both graph structure and feature information. With these estimated distributions, the probabilistic objective is then empirically estimated with samples, which leads to the enhanced objective. To evaluate the effectiveness of the proposed enhanced objective, we equipped it with a representative graph contrastive learning framework \gra~\cite{zhu2020deep} and its variant \gca ~\cite{zhu2021graph}. We also utilize it to help improve the performance of \gmlp~\cite{graph_mlp_cite}, a framework for training MLP with an auxiliary contrastive objective. Comprehensive experiment results suggest the enhanced objective can effectively improve the quality of contrast-based graph representation learning. 

%% file: sections/methodology.tex
\section{Preliminary}\label{sec:preliminary}
In this section, we introduce some basic notations and important concepts, which prepare us for further discussion in later sections. Specifically, we first introduce graphs and then briefly describe graph contrastive learning.

 Let $\mathcal{G}=\{\mathcal{V}, \mathcal{E}\}$ denote a graph with $\mathcal{V}$ and $\mathcal{E}$ denoting its set of nodes and edges, respectively. The edges describe the connections between the nodes, which can be summarized in an adjacency matrix ${\bf A}\in \{0,1\}^{|\mathcal{V}|\times |\mathcal{V}|}$ with $|\mathcal{V}|$ denoting the number of nodes. The $i,j$-th element of the adjacency matrix is denoted as ${\bf A}[i,j]$. It equals $1$ only when the nodes $i$ and $j$ connect to each other, otherwise $0$. Each node $i\in \mathcal{V}$ is associated with a feature vector ${\bf x}_i$.

Graph Contrastive Learning (GCL) aims to learn high-quality node representations by contrasting semantically similar and dissimilar node pairs. More specifically, given an anchor node $v\in \mathcal{V}$, those nodes with similar semantics as $v$ are considered positive samples, while those with dissimilar semantics are treated as negative samples. 
The goal of GCL is to pull the representations of those semantically similar nodes close and push the semantically dissimilar ones apart. From the perspective of a single anchor node $v$, the goal can be achieved by minimizing the following objective. 
{\small
\begin{align}
\mathcal{L}({v}) = -\log \frac{e^{f(v)^{\top} f\left(v^{{\prime}}\right)/\tau}}{ e^{f(v)^{\top} f\left(v^{{\prime}}\right)/\tau} + \sum_{v_s \in \mathcal{N}(v)} e^{f(v)^{\top} f\left(v_{s}\right)/\tau} },
\label{eqn:general_contrastive_loss}
\end{align}
}where $\tau$ is the temperature hyper-parameter, $v'$ is the positive sample, which is typically generated by data augmentation, $f(\cdot)$ is a function that maps a node $v$ to its low-dimensional representation, and $\mathcal{N}(v)$ denotes the negative samples corresponding to the anchor $v$. The overall objective for all nodes is a summation of $\mathcal{L}({v})$ over all nodes in $\mathcal{V}$. Next, we briefly introduce the positive sample, the $f(\cdot)$ function, and the set of negative samples as follows. 
\begin{itemize}
    \item Graph data augmentation techniques are typically utilized to create positive samples. Specifically, given a graph $\mathcal{G}$, an augmented graph $\mathcal{G}'$ is generated by perturbing $\mathcal{G}$. Then, for an anchor node $v$ in $\mathcal{G}$, its counterpart $v'$ in $\mathcal{G}'$ is treated its positive sample. Noticeably, $v^{{\prime}}$ is recognized as the only positive sample for $v$. Commonly adopted graph augmentation operations can be classified into two major categories: topology transformation and feature transformation \cite{zhu2020deep, you2020graph, zhu2021empirical}. 
    \item The function $f(\cdot)$ is a composition of an encoder $e(\cdot): \mathcal{V} \rightarrow \mathbb{R}^{d_1} $ and a projector $p(\cdot): \mathbb{R}^{d_1} \rightarrow \mathbb{R}^{d_2}$. Specifically, it can be formulated as $f(\cdot) = p(e(\cdot)) $. 
    The encoder aims to map a node to a representation vector in $d_1$ dimension $\ell_{2}$ normalized latent space. On the other hand, the projector aims to further map the representations output from the encoder to a $\mathbb{R}^{d_2}$ space specifically prepared for GCL. Note that, after GCL, the representations from the encoder will be utilized for downstream tasks. GNNs are often adopted as the encoder for GCL~\cite{velickovic2019deep, hassani2020contrastive}. 
    \item As we do not have access to the true semantics of nodes, the set of a negative sample $\mathcal{N}(v)$ is often uniformly sampled from the entire set of nodes $\mathcal{V}$. When dealing with large-scale graphs. batch-wise training is typically adopted ~\cite{hafidi2022negative}. In such a scenario, other nodes in the same batch as the anchor node are treated as negative samples~\cite{hafidi2022negative}.
\end{itemize}
In many existing GCL frameworks~\cite{zhu2020deep, zhu2021graph}, two augmented graphs $\mathcal{G}'$ and $\mathcal{G}''$ are typically generated and all nodes in these two graphs will be treated as anchor nodes. These two graphs are often named as two views of the original graph $\mathcal{G}$. In this case, the positive sample of a given anchor node is its counterpart in another view. The set of negative samples contains nodes from both views. \gra is such a representative GCL framework~\cite{zhu2020deep} for node-level tasks. It adopts edge removing and feature masking for conducting graph data augmentation. \gca \cite{zhu2021graph}, as an advanced variant of \gra, adopts an adaptive augmentation approach that is more likely to perturb unimportant edges and features.

\section{Methodology}\label{sec:methodology}
Though the objective described in Eq.~\eqref{eqn:general_contrastive_loss} has been widely adopted and has led to strong performance, it still suffers from several limitations. In particular, it only exploits one positive sample per anchor node, which limits its ability to learn high-quality representations. Furthermore, the uniform negative sampling strategy inevitably includes ``false-negative'' samples with similar semantics as negative samples, which further deteriorates the quality of the representations. In this work, we propose an enhanced objective to address these issues. We start our discussion with an ideal objective (see Section~\ref{sec:ideal}) that includes more positive samples for the numerator of Eq.~\eqref{eqn:general_contrastive_loss} and zero ``false-negative sample'' in the denominator. We then proceed to make the objective more practical by proposing an efficient strategy to empirically estimate the ideal objective (see details in Section~\ref{sec:estimation_of_objective}). Specifically, the estimation requires modeling the anchor-aware distributions for sampling positive samples and negative samples. Intuitively, those nodes with higher similarity to the anchor node should have a higher probability to be sampled as positive samples while being less likely to be selected as negative samples. Therefore, following this intuition, we use pairwise node similarity to model anchor-aware distributions (see details in Section~\ref{sec:modeling_distributions}). Finally, we present and discuss the proposed enhanced objective in Section~\ref{sec:enhanced_objective}.

\subsection{The Ideal Objective for GCL}\label{sec:ideal}
To address the limitations of the conventional objective in Eq.~\eqref{eqn:general_contrastive_loss}, an ideal objective that enjoys the capability of learning high-quality representations would include all positive nodes in the numerator while only including true negative samples in the denominator. More specifically, such an ideal objective for an anchor node $v$ could be formulated as follows. 
{\small
\begin{align}
    \mathcal{L}_{ideal}(v) = -\log \frac{ 
\sum_{v_j \in \mathcal{V}} \mathbbm{1}_{\left[y=y_{j}\right]} e^{f\left(v\right)^{\top} f\left(v_{j}\right) / \tau}
 }{e^{f\left(v\right)^{\top} f\left(v'\right) / \tau} +  \sum_{v_j \in \mathcal{V}} \mathbbm{1}_{\left[y\not=y_{j}\right]} e^{f\left(v\right)^{\top} f\left(v_{j}\right) / \tau} }, \label{eqn:multi_contrastive_loss}
\end{align}
}\noindent where $y$ denotes the ground truth label for node $v$ and $y_j$ is the label of $v_j$, and $\mathbbm{1}_{[a]}$ is an indicator function, which outputs $1$ if and only if the argument $a$ holds true, otherwise $0$.   

Nevertheless, the objective function in Eq.~\eqref{eqn:multi_contrastive_loss} is not achievable, as it is impossible to know the semantic classes of the downstream tasks in the contrastive training process, let alone the ground-truth labels. Hence, to make the objective more practical, in this paper, following the assumptions in~\cite{arora2019theoretical, chuang2020debiased, robinson2020contrastive}, we assume there are a set of discrete latent classes $\mathcal{C}$ standing for the true semantics of each node. We use $h:\mathcal{V}\rightarrow \mathcal{C}$ to denote the function mapping a given node to its latent class. For a node $v\in \mathcal{V}$, $h(v)$ denotes its latent class. Then, we introduce two types of anchor-aware sampling distributions over the entire node set $\mathcal{V}$. Specifically, for an anchor node $v$, we denote the probability of observing any node $u$ sharing the same latent class (i.e., $u$ is a positive sample corresponding to $v$) as $p_{v}^{+}\left(u \right)= p\left(u \mid h\left(u\right)=h(v)\right)$. Similarly, $p_{v}^{-}\left(u \right)= p\left(u \mid h\left(u\right)\not=h(v)\right)$ denotes the probability of observing $u$ as a negative sample corresponding to $v$. Note that the subscript in $p_v^+$ and $p_v^-$ indicates that they are specific to the anchor node $v$. With these two types of distributions, we can estimate the objective in Eq.~\eqref{eqn:multi_contrastive_loss} with positive and negative nodes sampled from the two distributions. Specifically, we estimate the objective as follows. 

{\small
\begin{align}
    \mathcal{L}_{est}(v) = \mathbb{E}_{\substack{\{v_{j}^{+}\}_{j=1}^m \sim p_{v}^{+}, \\ \{v_{k}^{-} \}_{k=1}^n \sim p_{v}^{-}}} 
 \left[   -\log \frac{ 
\frac{l}{m}\sum_{j=1}^m e^{f\left(v\right)^{\top} f\left(v^+_{j}\right) / \tau}
 }{e^{f\left(v\right)^{\top} f\left(v'\right) / \tau} + \frac{q}{n} \sum_{k}^n  e^{f\left(v\right)^{\top} f\left(v^-_{k}\right) / \tau} } \right], \label{eqn:estimate_ideal_contrastive_loss}
\end{align}
}
\noindent where $\{v_j^+\}_{j=1}^m$ and $\{v_k^-\}_{k=1}^n$ denote the set of ``positive nodes'' and ``negative nodes'' sampled following $p^+_v$ and $p^-_v$, respectively; and $m$ and $n$ denotes the number positive and negative samples, respectively. As similar to ~\cite{chuang2020debiased}, for the purpose of asymptotic analysis, we introduce two weight parameters $l$ and $q$. When $m$ and $n$ are finite, we set $l=n$ and $q=m$, which ensures that Eq.~\eqref{eqn:estimate_ideal_contrastive_loss} follows the same form as Eq.~\eqref{eqn:multi_contrastive_loss}.  

While the objective in Eq.~\eqref{eqn:estimate_ideal_contrastive_loss} is more practical than the one in Eq.~\eqref{eqn:multi_contrastive_loss}, it is still not achievable due to the following reasons: 1) we do not have access to the two anchor-aware distributions $p^+_v$ and $p^-_v$; and 2) even if we know these two distributions, the sampling complexity will still be high to provide an accurate estimate for the expectation. Next, we aim to address these two challenges. For the first challenge, we propose to model anchor-aware distributions utilizing both the graph structure and the feature information. Since directly modeling the distribution over all nodes in the graph is extremely difficult, we propose to connect the probabilities $p^+_v(u)$ and $p^-_v(u)$ of a specific node $u$ with node similarity between node $u$ and the anchor node $v$. Then, we utilize both the graph structure and feature information to model such a node similarity. More details on modeling anchor-aware distributions will be discussed in Section~\ref{sec:modeling_distributions}. For addressing the second challenge, we perform an asymptotic analysis, which leads to a new objective requiring fewer samples for estimation. The details of the asymptotic analysis and the new objective will be discussed in Section~\ref{sec:estimation_of_objective}. Next, we first discuss how we address the second challenge assuming we are given the two sets of anchor-aware distributions $p^+_v$ and $p^-_v$ in Section~\ref{sec:estimation_of_objective} and then discuss how we model the anchor-ware distributions $p^+_v$ and $p^-_v$ in Section~\ref{sec:modeling_distributions}. The proposed enhanced objective will be discussed in Section~\ref{sec:enhanced_objective}.

\subsection{Efficient Estimation of the Ideal Objective}\label{sec:estimation_of_objective}

To allow a more efficient estimation of Eq.~\eqref{eqn:estimate_ideal_contrastive_loss}, we consider its asymptotic form by analyzing the case where $m$ and $n$ go to infinity, which is summarized in the following theorem. 

\begin{theorem}\label{the:asymptotic}
\emph{For fixed $l$ and $q$, when $m \rightarrow \infty$ and $n \rightarrow \infty$,  it holds that}:

{\small
\begin{align*}
\mathbb{E}_{\substack{ \{v_{j}^{+}\}_{j=1}^{m} \sim p_{v}^{+} \\ \left\{v_{k}^{-}\right\}_{k=1}^{n} \sim p_{v}^{-}}}\left[-\log \frac{ \frac{l}{m} \sum_{j=1}^{m} e^{f(v)^{\top} f(v^{+}_{j})/\tau}}{ e^{f(v)^{\top} f(v^{\prime})/\tau}+\frac{q}{n} \sum_{k=1}^{n} e^{f(v)^{\top} f\left(v_{k}^{-}\right)/\tau}}\right]
\end{align*}
}

{\small
\begin{align*}
\rightarrow  -\log  \frac{ l\mathbb{E}_{ {v}^{+} \sim p_{v}^{+}({v}^{+})} [e^{f(v)^{\top} f\left({v}^{+}\right)/\tau}]}{   e^{f(v)^{\top} f(v^{\prime})/\tau}+ q\mathbb{E}_{{v}^{-} \sim p_{v}^{-}({v}^{-})} [e^{f(v)^{\top} f\left({v}^{-}\right)/\tau}]}.
\end{align*}
}

\begin{proof}
As $\tau$ is a nonzero scalar, the contrastive objective is bounded. Thus, we could apply the Dominated Convergence Theorem to prove the theorem above as follows:

{\small
\begin{align}
    &\lim _{m \rightarrow \infty} \lim _{n \rightarrow \infty} \mathbb{E}\left[-\log  \frac{  \frac{l}{m} \sum_{j=1}^{m} e^{f(v)^{\top} f(v^{+}_{j})/\tau}}{ e^{f(v)^{\top} f(v^{\prime})/\tau}+\frac{q}{n} \sum_{k=1}^{n} e^{f(v)^{\top} f\left(v_{k}^{-}\right)/\tau}}\right] \nonumber \\
    =& \mathbb{E}\left[ \lim _{m \rightarrow \infty} \lim _{n \rightarrow \infty} -\log  \frac{  \frac{l}{m} \sum_{j=1}^{m} e^{f(v)^{\top} f(v^{+}_{j})/\tau}}{ e^{f(v)^{\top} f(v^{\prime})/\tau}+\frac{q}{n} \sum_{k=1}^{n} e^{f(v)^{\top} f\left(v_{k}^{-}\right)/\tau}}\right] \nonumber\\
    =&  \mathbb{E}\left[ -\log  \frac{ l\mathbb{E}_{ {v}^{+} \sim p_{v}^{+}({v}^{+})} [e^{f(v)^{\top} f\left({v}^{+}\right)/\tau}]}{   e^{f(v)^{\top} f(v^{\prime})/\tau}+ q\mathbb{E}_{{v}^{-} \sim p_{v}^{-} ({v}^{-})} [e^{f(v)^{\top} f\left({v}^{-}\right)/\tau}]}\right]\nonumber\\
    =&  -\log  \frac{ l\mathbb{E}_{ {v}^{+} \sim p_{v}^{+}({v}^{+})} [e^{f(v)^{\top} f\left({v}^{+}\right)/\tau}]}{   e^{f(v)^{\top} f(v^{\prime})/\tau}+ q\mathbb{E}_{{v}^{-} \sim p_{v}^{-} ({v}^{-})} [e^{f(v)^{\top} f\left({v}^{-}\right)/\tau}]}.
 \label{eqn:asy_contrastive_loss} 
\end{align}
}
\end{proof}
\end{theorem}

As demonstrated in Theorem~\ref{the:asymptotic}, the objective of Eq.~\eqref{eqn:asy_contrastive_loss} is an asymptotic form of Eq.~\eqref{eqn:estimate_ideal_contrastive_loss}. In this work, we aim to empirically estimate Eq.~\eqref{eqn:asy_contrastive_loss} instead of Eq.~\eqref{eqn:estimate_ideal_contrastive_loss}. Specifically, Eq.~\eqref{eqn:asy_contrastive_loss} contains two expectations to be estimated. Compared to Eq.~\eqref{eqn:estimate_ideal_contrastive_loss}, the sampling complexity is significantly reduced, as we disentangled the joint distribution in Eq.~\eqref{eqn:estimate_ideal_contrastive_loss}, and only need to estimate these two expectations independently. More specifically, to estimate $\mathbb{E}_{ {v}^{+} \sim p_{v}^{+}({v}^{+})} [e^{f(v)^{\top} f\left({v}^{+}\right)/\tau}]$, a straightforward way is to randomly draw samples from $p_{v}^{+}$ and calculate its empirical mean. However, it is typically inefficient and inconvenient to obtain samples directly from $p_{v}^{+}$, as $p_{v}^{+}$ itself needs to be estimated (this will be discussed in Section~\ref{sec:modeling_distributions}) and we cannot obtain a simple analytical form to perform the sampling. The same reason applies to the estimation of $\mathbb{E}_{{v}^{-} \sim p_{v}^{-} ({v}^{-})} [e^{f(v)^{\top} f\left({v}^{-}\right)/\tau}]$. 
Therefore, in this work, we adopt the importance sampling strategy~\cite{Goodfellow-et-al-2016} to estimate the two expectations using samples from the uniform distribution $p$ as follows.
{\small
\begin{align}
    \mathbb{E}_{ {v}^{+} \sim p_{v}^{+}} [e^{f(v)^{\top} f\left({v}^{+}\right)/\tau}] &=  \mathbb{E}_{ {v}^{+} \sim p} \left[\frac{p^+_v(v^+)}{p(v^+)}e^{f(v)^{\top} f\left({v}^{+}\right)/\tau}\right] \nonumber\\
    & \approx \frac{1}{M}\sum\limits_{v_j \in \mathcal{V}_M} \left[\frac{p^+_v(v_j)}{p(v_j)}e^{f(v)^{\top} f\left({v}_j\right)/\tau}\right]; \label{eqn:importance_sample_form_pos}\\
  \mathbb{E}_{ {v}^{-} \sim p_{v}^{-}} [e^{f(v)^{\top} f\left({v}^{-}\right)/\tau}] &=  \mathbb{E}_{ {v}^{-} \sim p} \left[\frac{p^-_v(v^-)}{p(v^-)}e^{f(v)^{\top} f\left({v}^{-}\right)/\tau}\right] \nonumber\\
  & \approx \frac{1}{N}\sum\limits_{v_j \in \mathcal{V}_N} \left[\frac{p^-_v(v_j)}{p(v_j)}e^{f(v)^{\top} f\left({v}_j\right)/\tau}\right], 
  \label{eqn:importance_sample_form} 
\end{align}
}
\noindent where $\mathcal{V}_M = {\{v_j\}}_{j=1}^M \sim p$ contains $M$ nodes sampled from $p$ and $\mathcal{V}_N = {\{v_j\}}_{j=1}^N \sim p$ contains $N$ nodes sampled from $p$, which are utilized for estimation. To obtain the final empirical form of Eq.~\eqref{eqn:asy_contrastive_loss},  the two sets of anchor-aware distributions $p^+_v$ and $p^-_v$ remain to be estimated, which is discussed in the next section.

\subsection{Modeling and Estimating Anchor-aware Distributions}\label{sec:modeling_distributions}
In this section, we discuss the modeling details of the anchor-aware distributions $p^+_v$ and $p^-_v$. As discussed earlier in Section~\ref{sec:ideal}, for an anchor $v$, the positive sample distribution is a conditional distribution relying on the agreement of the latent classes of $v$ and any other sample $u$, which can be formulated as $p_{v}^{+} (u) = p_v(u| h(v) = h(u))$. Direct modeling this distribution is impossible, since we do not have access to the latent semantic class. In this section, we propose to model $p_{v}^{+}(u)$ with the node similarity between the anchor node $v$ and a given sample $u$ (Section~\ref{sec:modeling_distributions} and Section~\ref{sec:transformation}). We then discuss the process to evaluate node similarity with both graph structure and node feature information in Section~\ref{sec:similarity}.  

\subsubsection{Modeling anchor-aware distributions with node similarity.}
Based on Bayes' Theorem, we have
{\small
\begin{align}
    p_{v}^{+}(u)  \propto p_v(h(v) = h(u)| u)p(u),\label{eqn:propto_condition}
\end{align}
}\noindent where $p$ is a uniform distribution over all nodes, and $p_v(h(v) = h(u)$ is the probability that $u$ shares the same latent semantic class of $v$. Therefore, to obtain $ p_{v}^{+}(u)$, it is essential to model $p_v(h(v) = h(u)| u)$ as $p$ is already known. Intuitively, if $v$ and $u$ are more ``similar'' to each other, they are more likely to share the same semantic class. Assuming that we are given a function $\text{sim}(\cdot, \cdot)$ that measures the pair-wise similarity of any two nodes, then we further assume that the probability  $p_v(h(v) = h(u)| u)$ is positively correlated with $\text{sim}(v,u)$, which can be formulated as 
{\small
\begin{align}
    p_v(h(v) = h(u)| u) \propto \mathcal{T}(\text{sim}(v,u)), \label{eqn:propto_sim}
\end{align}
}\noindent where $\mathcal{T}$ is a monotonic increasing transformation. We will discuss the details of the transformation and the similarity function in Section~\ref{sec:transformation} and Section~\ref{sec:similarity}, respectively. Together with Eq.~\eqref{eqn:propto_condition}, we have 
{\small
\begin{align}
    p_{v}^{+}(u)  \propto \mathcal{T}(\text{sim}(v,u))p(u), \label{eqn:propto_with_sim}
\end{align}
}which intuitively expresses that those samples that are more similar to $v$ are more likely to be sampled as positive samples. We then formulate the probability $p_{v}^{+}(u)$ with $\text{sim}(v,u)$ as follows.
{\small
\begin{align}
&p_{v}^{+} (u) = \frac{ \mathcal{T}(\text{sim}(v,u)) p(u)}{\int \mathcal{T}(\text{sim}(v,v^{s})) p({v}^{s})d{v}^{s}} =  \frac{\mathcal{T}(\text{sim}(v,u)) p(u)}{\mathbb{E}_{v^s \sim p} [\mathcal{T}(\text{sim}(v,v^s))]}.
\label{eqn:pos_prop_with_sim} 
\end{align}
}Note that, in practice, $\mathbb{E}_{v^s \sim p} [\mathcal{T}(\text{sim}(v,v^s))]$ can be empirically estimated using the set of samples $\mathcal{V}_M$ in Eq.~\eqref{eqn:importance_sample_form_pos} as follows.
{\small
\begin{align}
    \mathbb{E}_{v^s \sim p} [\mathcal{T}(\text{sim}(v,v^s))] \approx \frac{1}{M} \sum\limits_{{v_j\in \mathcal{V}_M}} \mathcal{T}(\text{sim}(v,v_j)). 
\end{align}
}
Then, we can estimate $p^+(u)$ as follows
{\small
\begin{align}
&\hat{p}_{v}^{+} (u) = \frac{\mathcal{T}(\text{sim}(v,u)) p(u)}{\frac{1}{M} \sum\limits_{{v_j \in \mathcal{V}_M} } \mathcal{T}(\text{sim}(v,v_j))},
\label{eqn:pos_estimation_with_sim} 
\end{align}
}\noindent where $\hat{p}_{v}^{+} (u)$ is the empirical estimate of $p_{v}^{+} (u)$. Intuitively, given $\hat{p}_{v}^{+} (u)$, we can directly estimate $\hat{p}_{v}^{-} (u)$ as $1 - \hat{p}_{v}^{+} (u)$. However, this is typically not optimal for the purpose of contrastive learning for several reasons: 1) first, the samples in $\mathcal{V}_M$( in Eq.~\eqref{eqn:importance_sample_form_pos}) and $\mathcal{V}_N$ (in Eq.~\eqref{eqn:importance_sample_form}) are likely different, which makes it infeasible to directly model $p_v^-$ using $1-p_v^+$ for all selected nodes; 2) second, we prefer different properties of the estimations for the two distributions $p_v^+$ and $p_v^-$ for the purpose of contrastive learning. Specifically, we prefer a relatively conservative estimation for $p_v^+$ to reduce the impact of ``false positives'' (i.e, avoid assigning high $p^+_v$ for real negative samples). In contrast, a more aggressive estimation of $p_v^-$ is acceptable. Modeling a conservative  $p_v^+$ and aggressive $p_v^-$ at the same time cannot be achieved if we constrain $p_v^+(u) + p_v^-(u)=1$. Due to the above reasons, in this work, we relax this constraint and model $p_v^-$ flexibly using node similarity $\text{sim}(\cdot, \cdot)$ as follows. 
{\small
\begin{align}
&p_{v}^{-} (u) =   \frac{\mathcal{D}(\text{sim}(v,u)) p(u)}{\mathbb{E}_{v^s \sim p} [\mathcal{D}(\text{sim}(v,v^s))]},
\label{eqn:neg_prop_with_sim} 
\end{align}
}\noindent where $\mathcal{D}$ is a monotonic decreasing function, indicating that $p_v^-$ is negatively correlated with the similarity. Similar to Eq.~\eqref{eqn:pos_estimation_with_sim}, $p_{v}^{-} (u)$ can be empirically estimated with $N$ samples in $\mathcal{V}_N$ (described in Eq.~\eqref{eqn:importance_sample_form}) as follows. 
{\small
\begin{align}
&\hat{p}_{v}^{-} (u) = \frac{\mathcal{D}(\text{sim}(v,u)) p(u)}{\frac{1}{N} \sum\limits_{v_j \in \mathcal{N}} \mathcal{D}(\text{sim}(v,v_j))}.
\label{eqn:neg_estimation_with_sim} 
\end{align}
}Next, we first discuss the details of the monotonic increasing transformation function $\mathcal{T}$ and the monotonic decreasing transformation $\mathcal{D}$ in Section~\ref{sec:transformation}. We then discuss the similarity function $\text{sim}(v,u)$ in Section~\ref{sec:similarity}. 

\subsubsection{Transformations}\label{sec:transformation}
To flexibly adjust the two estimated anchor-aware distributions $\hat{p}^+_v$ and $\hat{p}_v^-$ between conservative estimation to aggressive estimation, we utilize exponential function with temperature \cite{doi:10.1080/002068970500044749} to model the transformation functions as follows
{\small
\begin{align}
    &\mathcal{T}(\text{sim}(v_i,v_j)) = \exp(\text{sim}(v_i,v_j)/\tau_p)-1;\label{eqn:positive_transformation}\\
    &\mathcal{D}(\text{sim}(v_i,v_j)) = \exp(-\text{sim}(v_i,v_j)/\tau_n), \label{eqn:negative_transformation}
\end{align} 
}\noindent where $\tau_p$ and $\tau_n$ are two temperature parameters. We could adjust the estimation of the two distributions $\hat{p}_v^+$ in Eq.~\eqref{eqn:pos_estimation_with_sim} and $\hat{p}_v^-$ in Eq.~\eqref{eqn:neg_estimation_with_sim} by varying $\tau_p$ and $\tau_n$, respectively. More specifically, for $\hat{p}_v^+$, we could make the distribution more conservative by decreasing $\tau_p$, which increases the probability mass for those samples with high similarity. In the extreme case, when $\tau_p$ goes to $0$, the probability mass concentrates in the sample with the largest similarity. On the other hand, when $\tau_p$ reaches infinity, $\hat{p}_v^+$ converges to a distribution proportional to similarity. Note that without the ``$-1$'' in Eq.~\eqref{eqn:positive_transformation}, $\hat{p}_v^+$ converges to a uniform distribution as $\tau_p$ goes to infinity, which leads to model collapse as all samples in Eq.~\eqref{eqn:importance_sample_form_pos} will be treated equally (all treated as positive samples). Thus, we include ``$-1$'' in Eq.~\eqref{eqn:positive_transformation} to avoid such cases. Similarly, $\hat{p}_v^-$ can be adjusted from a uniform distribution to a distribution with mass concentrated on the sample with the smallest similarity by varying $\tau_n$. Specifically, when $\tau_n$ goes to $0$, the estimated $\hat{p}_v^-$ converges to the uniform distribution and the estimation in Eq.~\eqref{eqn:importance_sample_form} reduces to the same result as the convectional negative sampling strategy.  

\subsubsection{Modeling node similarity.} \label{sec:similarity}
To comprehensively evaluate the similarity between nodes, it is of great importance to capture the similarity in terms of both graph structure and node features. In this section, we aim to model the overall similarity function by estimating and combining the structure similarity and feature similarity. Next, we first describe how we model these two types of similarity individually and then discuss how we combine them to model the overall similarity function. 

\noindent\emph{\bf Graph Structure Similarity} Personalized Page Rank (PPR) is a widely adopted tool for measuring the relevance between nodes in graph mining~\cite{page1999pagerank, park2019survey, lamurias2019ppr}. More recently, it has also been adopted to improve graph representation learning~\cite{klicpera2018predict, klicpera2019diffusion}. Therefore, in this work, we utilize the PPR score to model the structural node similarity. Specifically, the personalized PageRank matrix is defined as ${\bf P} = \alpha({\bf I}-(1-\alpha)\hat{\bf A})^{-1}$, where $\hat{\bf A}={ \bf D}^{-1/2} {\bf A} {\bf D}^{-1 / 2}$, ${\bf D}$ is the degree matrix, and $\alpha\in(0,1)$ is a hyper-parameter. The $i,j$-th element of the PPR matrix ${\bf P}$ denoting as ${\bf P}[i,j]$ measures the structural similarity between node $v_i$ and node $v_j$. However, calculating the matrix ${\bf P}$ is computationally expensive, especially for large-scale graphs, as it involves a matrix inverse. In this work, we adopt the iterative approximation of the PPR matrix for measuring the node similarity as $
    \hat{\bf P} = (1-\alpha)^{K}\hat{\bf A}^{K} + \sum\limits_{k=0}^{K-1}  \alpha(1-\alpha)^k \hat{\bf A}^k,$ where $K$ is the number of iterations. Note that, $\hat{\bf P}$ converges to ${\bf P}$ as $K$ goes infinity~\cite{klicpera2018predict}. Based on $\hat{\bf P}$, we also investigate another higher-order structure similarity measure from a more global perspective. Specifically, for two given nodes $v_i$ and $v_j$, we utilize the cosine similarity of the $i$-th row and $j$-th row as the similarity between these two nodes. We evaluate the structure similarity between two nodes $v_i$ and $v_j$ either by $\text{sim}_G(v_i,v_j) = \hat{P}[i,j]$ or $\text{sim}_G(v_i,v_j) = \cos(\hat{P}[i,:], \hat{P}[j,:])$, where $\hat{P}[i,:]$ denotes the $i$-th row of the approximated PPR matrix $\hat{P}$. In the experiments, we treat the choice between these two types of similarity measure as a ``hyper-parameter'' to be tuned. There exist many other methods for measuring the structural node similarity, and we leave them for future work. 

\noindent\emph{\bf Feature Similarity.} To better mine the pairwise node similarity from features, we adopt the classic cosine similarity. Specifically, feature similarity between nodes $v_i$ and $v_j$ is evaluated by $\text{sim}_F(v_i,v_j) = \cos({\bf x}_i,{\bf x}_j)$, where ${\bf x}_i, {\bf x}_j$ are the original input features of node $v_i$ and $v_j$, respectively. 

\noindent\emph{\bf Fusing Graph and Feature Similarity.} 
Given the structure similarity $\text{sim}_G(v_i,v_j)$ and feature similarity $\text{sim}_F(v_i,v_j)$, it is vital to define an adaptive function $\text{sim}(\cdot, \cdot)$ to fuse them and output a combined similarity score capturing information from both sources. Specifically, we propose to combine the two similarities to form the overall similarity as 
$
\text{sim}(v_i,v_j)= \beta \cdot \text{sim}_F(v_i,v_j)\cdot \gamma + (1-\beta)\cdot  \text{sim}_G(v_i,v_j),
$
where $\gamma$ is the scaling factor to control the relative scale between the two similarity scores, and $\beta$ is a hyper-parameter balancing the two types of similarity. In general, $\gamma$ could also be treated as a hyper-parameter. In this work, we fix $\gamma = \sum sim_G(v_i,v_j)/\sum sim_F(v_i,v_j)$ such that the two types of similarity are at the same scale. 

\subsection{The Proposed Enhanced Objective}\label{sec:enhanced_objective}
With th estimation of $\hat{p}^+_v$ in Eq.~\eqref{eqn:pos_estimation_with_sim} and $\hat{p}^-_v$ in Eq.~\eqref{eqn:neg_estimation_with_sim}, we propose an enhanced objective as follows.
{\small
\begin{align}
    \mathcal{L}_{EN}(v) =  -\log  \frac{\sum\limits_{v_j\in \mathcal{V}_M} \left[w^+_v(v_j)e^{f(v)^{\top} f\left({v}_j\right)/\tau}\right] }{   e^{f(v)^{\top} f(v^{\prime})/\tau}+  \sum\limits_{v_j\in \mathcal{V}_N} \left[w^-_v(v_j)e^{f(v)^{\top} f\left({v}_j\right)/\tau}\right] },
    \label{eqn:final_loss}
\end{align}
}
\noindent where $w^+_v(v_j)$ and $w^-_v(v_j)$ are defined as follows.
{\small
\begin{equation}
\begin{aligned}
    w^+_v(v_j) &=  \frac{\mathcal{T}(\text{sim}(v,v^j)) }{\frac{1}{M} \sum\limits_{v_j\in \mathcal{V}_M} \mathcal{T}(\text{sim}(v,v_k))};\\
      w^-_v(v_j) & = \frac{\mathcal{D}(\text{sim}(v,v_j))}{\frac{1}{N} \sum\limits_{v_j\in \mathcal{V}_N} \mathcal{D}(\text{sim}(v,v_k))}.
    \label{eqn:weight}
\end{aligned}
\end{equation}}$\mathcal{V}_M$ and $\mathcal{V}_N$ are the two sets of nodes introduced in Eq.~\eqref{eqn:importance_sample_form_pos} and Eq.~\eqref{eqn:importance_sample_form}. 
If we set $\mathcal{V}_M = \mathcal{V}_N=\mathcal{V}$, we can make a direct comparison between the enhanced objective in Eq.~\eqref{eqn:final_loss} and the ideal objective in Eq.~\eqref{eqn:multi_contrastive_loss}. Specifically, the enhanced objective can be considered as a soft version of the ideal objective described at Eq.~\eqref{eqn:multi_contrastive_loss}, where the weights $w^+_v(v_j)$ and $w^-_v(v_j)$ in Eq.~\eqref{eqn:final_loss} reflects the likelihood of $v_j$ being a positive sample or a negative sample, respectively.

\section{Beyond Graph Contrastive Learning}\label{sec:gmlp-intro}
The philosophy of contrastive learning has inspired other frameworks for graph representation learning. In \gmlp~\cite{graph_mlp_cite}, an auxiliary neighborhood contrastive loss is proposed to enhance the performance of MLP on the semi-supervised node classification task. As indicated by its name, the key idea of the neighborhood contrastive loss is to treat the ``neighboring nodes'' as positive samples and contrast them with their corresponding anchor nodes. Since the neighbors are defined through graph structure, such a loss helps incorporate the graph information into the representation learning process of MLP. It has been shown that the MLP model equipped with the neighborhood contrastive loss is capable of achieving performance comparable to or even stronger than graph neural network models. In this section, we briefly describe the \gmlp model with neighborhood contrastive loss and discuss how the proposed techniques discussed in Section~\ref{sec:methodology} can be utilized to further enhance this loss. 

\subsection{Neighborhood Contrastive Loss and \gmlp}
For an anchor node $v_i\in \mathcal{V}$, the neighborhood contrastive loss is defined as follows:
{\small
\begin{align}
\mathcal{L}_{NC}(v_i)=-\log \frac{\sum\limits_{v_j \in \mathcal{V}_b} \mathbbm{1}_{[v_j \neq v_i]} \hat{\bf A}^r[i,j] e^{ f(v_i)^{\top} f(v_j)/\tau}}{\sum\limits_{v_k \in \mathcal{V}_b} \mathbbm{1}_{[v_k \neq v_i]} e^{ f(v_i)^{\top} f(v_k)/\tau} }, \label{eqn:individual_NC_loss}
\end{align}
}
\noindent where $\mathcal{V}_b$ is a set of nodes uniformly sampled from $\mathcal{V}$, $\hat{\bf A}^r$ is the $r$-th power of the normalized adjacency matrix $\hat{\bf A}$. The $i,j$-th element $\hat{\bf A}^r[i,j]$ is only non-zero when node $v_j$ is within the $r$-hop neighborhood of node $v_i$, otherwise $\hat{\bf A}^r[i,j]=0$. Hence, in the numerator of Eq.~\eqref{eqn:individual_NC_loss}, only the $r$-hop neighbors are treated as positive samples. The denominator is similar to that in contrastive learning. Overall, the neighborhood contrastive loss for all nodes in the graph can be formulated as follows.
{\small
\begin{align}
   \mathcal{L}_{NC} = \sum\limits_{i\in \mathcal{V} } \mathcal{L}_{NC}(v_i). \label{eqn:nc_loss}
\end{align}
}
In \gmlp, the neighborhood contrastive loss is combined with the cross-entropy loss for conventional semi-supervised node classification as $\mathcal{L}_{train} = \mathcal{L}_{CE} + \alpha \mathcal{L}_{NC},$ where $\alpha>0$ is a hyper-parameter that balances the cross-entropy loss $\mathcal{L}_{CE}$ and the neighborhood contrastive loss $\mathcal{L}_{NC}$. When the graph is large, the neighborhood contrastive loss can be calculated in a batch-wise way, where $\mathcal{L}_{NC}$ can be calculated over a batch of nodes as $\mathcal{L}_{NC} =  \sum\limits_{v_i\in \mathcal{B} } \mathcal{L}_{NC}(v_i) $ with $\mathcal{B}$ denoting a specific sampled batch. Correspondingly, in this scenario, $\mathcal{V}_b$ in Eq.~\eqref{eqn:individual_NC_loss} can be replaced by $\mathcal{B}$.
\subsection{Enhanced Objective for \gmlp}
 
Following the same philosophy as in Section~\ref{sec:methodology}, we propose the following enhanced neighborhood contrastive loss. 
{\small
\begin{align}
\mathcal{L}_{EN-NC}(v_i)=-\log \frac{\sum\limits_{v_j \in \mathcal{V}_b} \mathbbm{1}_{[v_j \neq v_i]}w^+_{v_i}(v_j) e^{ f(v_i)^{\top} f(v_j)/\tau}}{\sum\limits_{v_k \in \mathcal{V}_b} \mathbbm{1}_{[v_k \neq v_i]} w^-_{v_i}(v_k) e^{ f(v_i)^{\top} f(v_k)/\tau} }, \label{eqn:individual_en_NC_loss}
\end{align}
}\noindent where $w^+_{v_i}(v_j)$ and $w^-_{v_i}(v_k)$ are the positive weight between nodes $v_i, v_j$ and negative weight between nodes $v_i, v_k$ as defined in Eq.~\eqref{eqn:weight}. We can replace $\mathcal{L}_{NC}(v_i)$ in Eq.~\eqref{eqn:nc_loss} with $\mathcal{L}_{EN-NC}(v_i)$ to form an enhanced training framework for MLP models. We name such a framework as \gmlpP. Its superiority is empirically verified in the experiments section (Section~\ref{sec:graph-mlp-results}).

%% file: sections/experiment.tex
 \begin{figure*}[!htp]%
 \captionsetup[subfloat]{captionskip=0mm}
     \centering
     \subfloat[\cora]{{\includegraphics[width=0.24\linewidth]{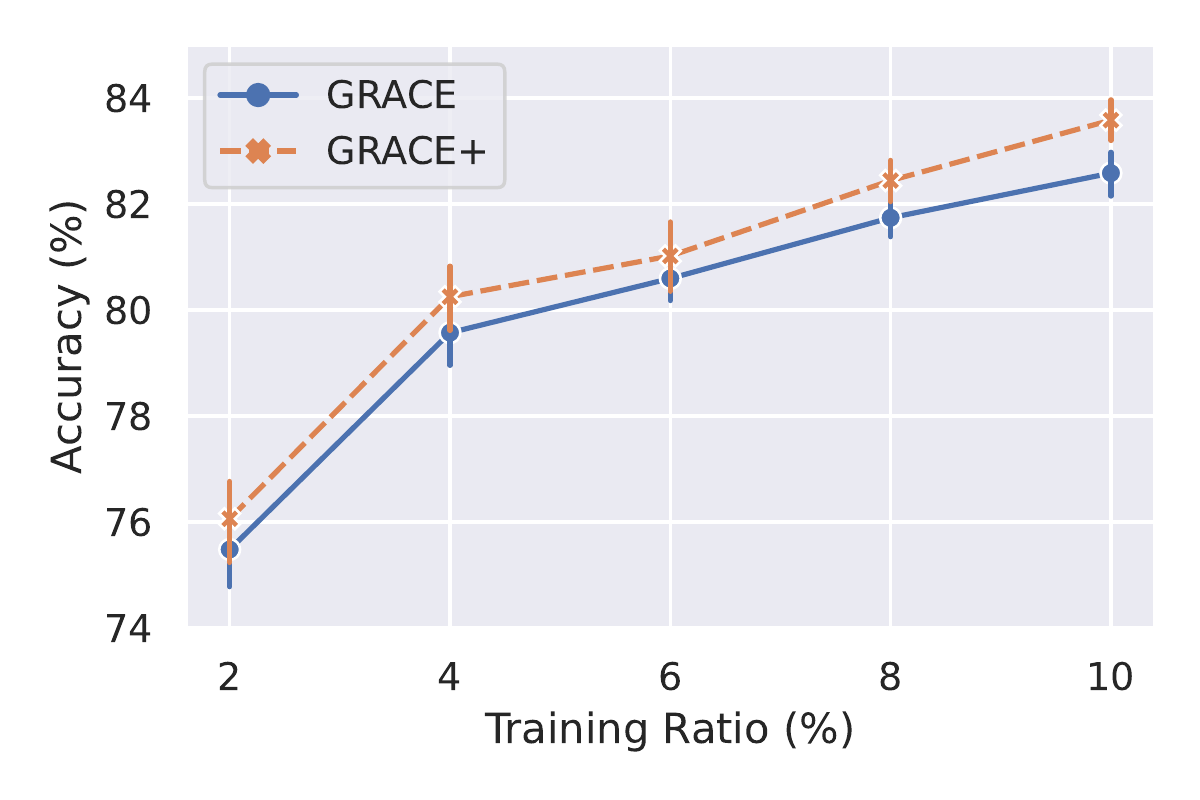}}}%
     \subfloat[\citeseer]{{\includegraphics[width=0.24\linewidth]{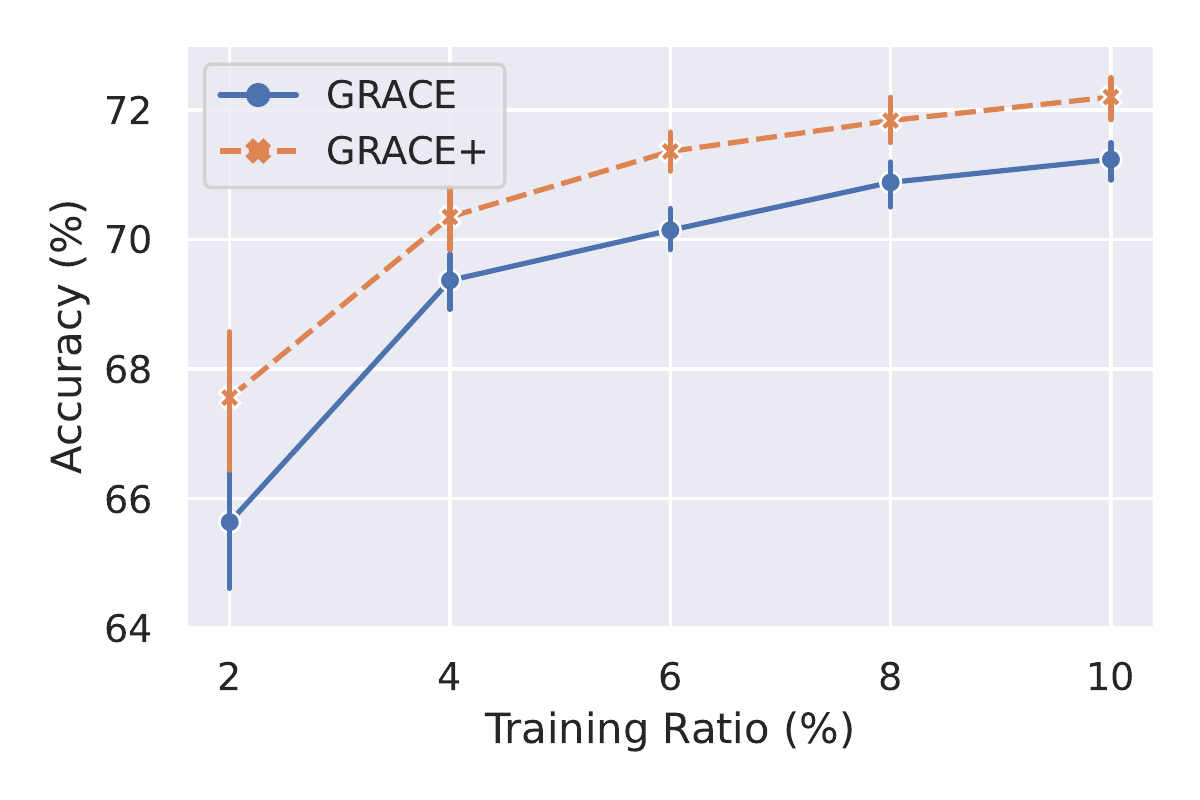} }}%
    \subfloat[\pubmed]{{\includegraphics[width=0.24\linewidth]{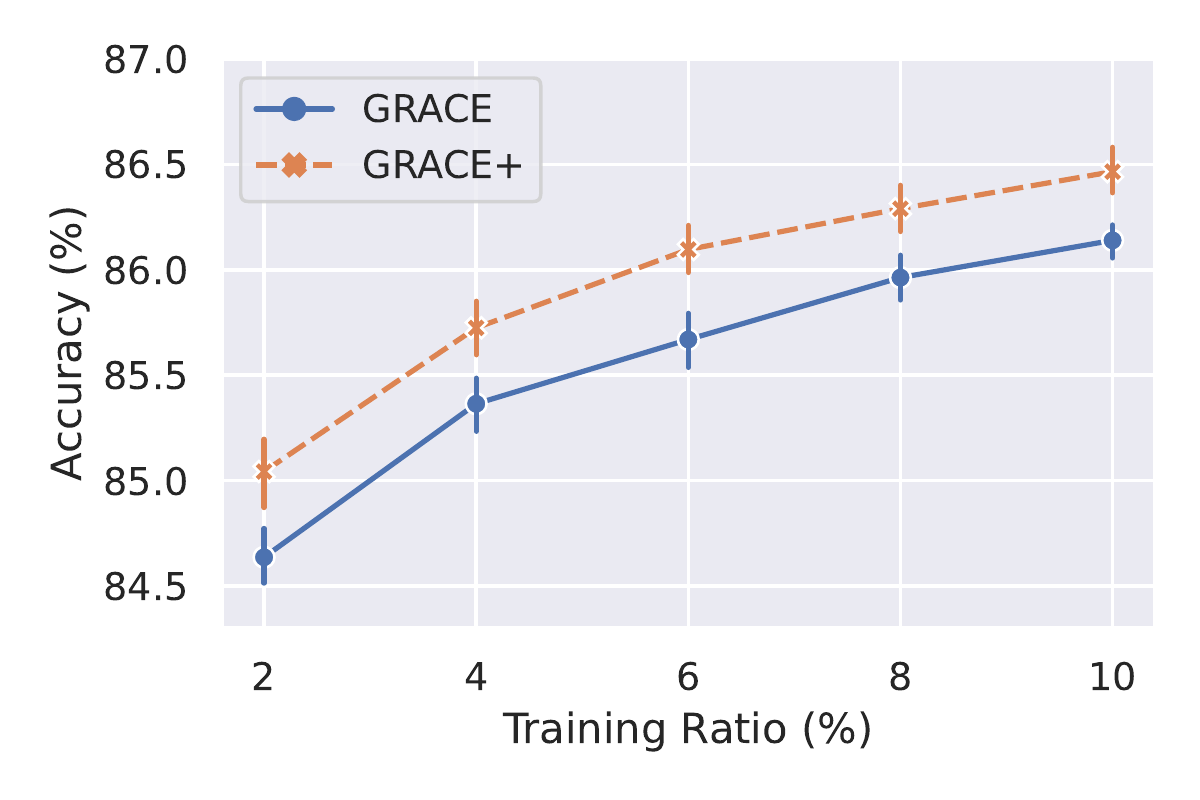} }}%
    \subfloat[\dblp]{{\includegraphics[width=0.24\linewidth]{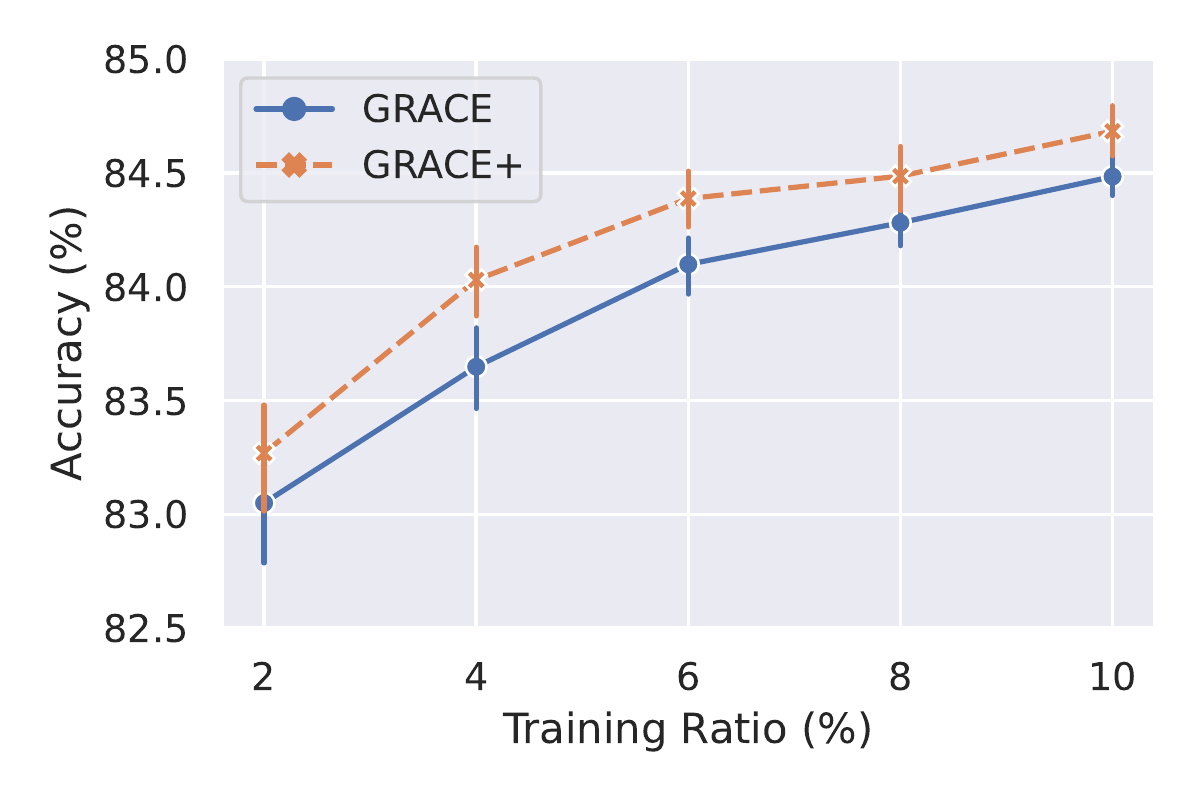} }}%
    \vspace{-3mm}
    \caption{\small{Node classification results of \gra and \graP with various training ratios.}} 
    \label{fig:grace-mlp-various-samples}
\end{figure*}
\section{Experiment}\label{sec:experiment}
In this section, we conduct experiments to verify the effectiveness of the enhanced objectives. Furthermore, we also perform an ablation study to provide a deep understanding of the proposed objectives. Next, we first introduce the datasets we adopt for experiments in Section~\ref{sec:dataset}. Then, we present the results with discussions for GCL and \gmlp in Section~\ref{sec:exp-gcl} and Section~\ref{sec:graph-mlp-results}, respectively. The ablation study is presented in Section~\ref{sec:ablation_study}. 

\subsection{Datasets}\label{sec:dataset}
In this section, we introduce the datasets we adopt for the experiments. Following previous papers~\cite{zhu2020deep,zhu2021graph,zhu2021graph}, we adopt $8$ datasets including \cora~\cite{sen2008collective}, \citeseer~\cite{sen2008collective}, \pubmed~\cite{sen2008collective}, \dblp~\cite{yang2015defining}, \photo ~\cite{shchur2018pitfalls}, \computer ~\cite{shchur2018pitfalls}, \cocs~\cite{shchur2018pitfalls}, and \wiki~\cite{mernyei2020wiki} for evaluating the performance. Some statistics of these datasets are in Table~\ref{tab:stats}.



\begin{table}[htb]
\caption{\small{Dataset Statistics}}
\vspace{-3mm}
\begin{adjustbox}{width=0.45\textwidth}
\begin{tabular}{ccccc}
\toprule
\textbf{Dataset} & \textbf{Nodes } & \textbf{Edges } & \textbf{Features Dim} & \textbf{Classes} \\ 
\midrule
\textbf{\cora}    & 2708  & 5429   & 1433 & 7 \\
\textbf{\citeseer} & 3327  & 4732   & 3703 & 6 \\ 
\textbf{\pubmed}   & 19717 & 44338  & 500  & 3 \\ 
\textbf{\dblp}     & 17716 & 105734 & 1639 & 4 \\ 
\midrule
\textbf{\computer}     & 13752 & 245861 & 767 & 10 \\ 
\textbf{\photo}     & 7650 & 119081 & 745 & 8 \\ 
\textbf{\wiki}     & 11701 & 216123 & 300 & 10 \\ 
\textbf{\cocs}     & 18333 & 81894& 6805 & 15 \\ 
 \bottomrule
\end{tabular}
\end{adjustbox}
\label{tab:stats}
 \end{table}

\subsection{Graph Contrastive Learning}\label{sec:exp-gcl}
Note that the enhanced objective proposed in Eq.~\eqref{eqn:final_loss} is quite flexible and can be utilized to improve the performance of various frameworks that adopt the conventional graph contrastive learning objective. In this work, we adopt \gra~\cite{zhu2020deep}, a recently proposed representative GCL framework and its updated version, \gca ~\cite{zhu2021graph}, as base models (check Section~\ref{sec:preliminary} for a brief description of \gra and \gca). We denote the \gra framework with the enhanced objective as \graP and use \gcaP to represent the enhanced \gca. Next, we first present results for \gra, and then describe results for \gca.
\subsubsection{\gra}\label{sec:grace} Following~\cite{zhu2020deep}, we conduct the experiment with \gra and \graP on the first $4$ citation datasets as introduced in Section~\ref{sec:dataset}. To evaluate the effectiveness of \graP, we adopt the same linear evaluation scheme as in~\cite{velickovic2019deep, zhu2020deep}. To be specific, the experiments are conducted in two stages. In the first stage, we learn node representations with the graph contrastive learning frameworks (\gra and \graP) in a self-supervised fashion. Then, in the second stage, we evaluate the quality of the learned node representations through the node classification task. Specifically, a logistic regression model with the obtained node representations as input is trained and tested. To comprehensively evaluate the quality of the representations, we adopt different training-validation-test splits. Specifically, we first randomly split the node sets into three parts: $80\%$ for testing, $10\%$ for validation, and the rest of $10\%$ is utilized to further build the training sets. With the remaining $10\%$ of nodes, we build $5$ different training sets that consist of $2\%, 4\%,6\%,8\%, 10\%$ of nodes in the entire graph. The training set is randomly sampled from the $10\%$ of data for building the training subset. We repeat the experiments $30$ times with different random initialization and report the average performance with standard deviation.

\begin{figure*}[htp]%
\vspace{-6mm}
 \captionsetup[subfloat]{captionskip=0mm}
     \centering
     \subfloat[Cora]{{\includegraphics[width=0.28\linewidth]{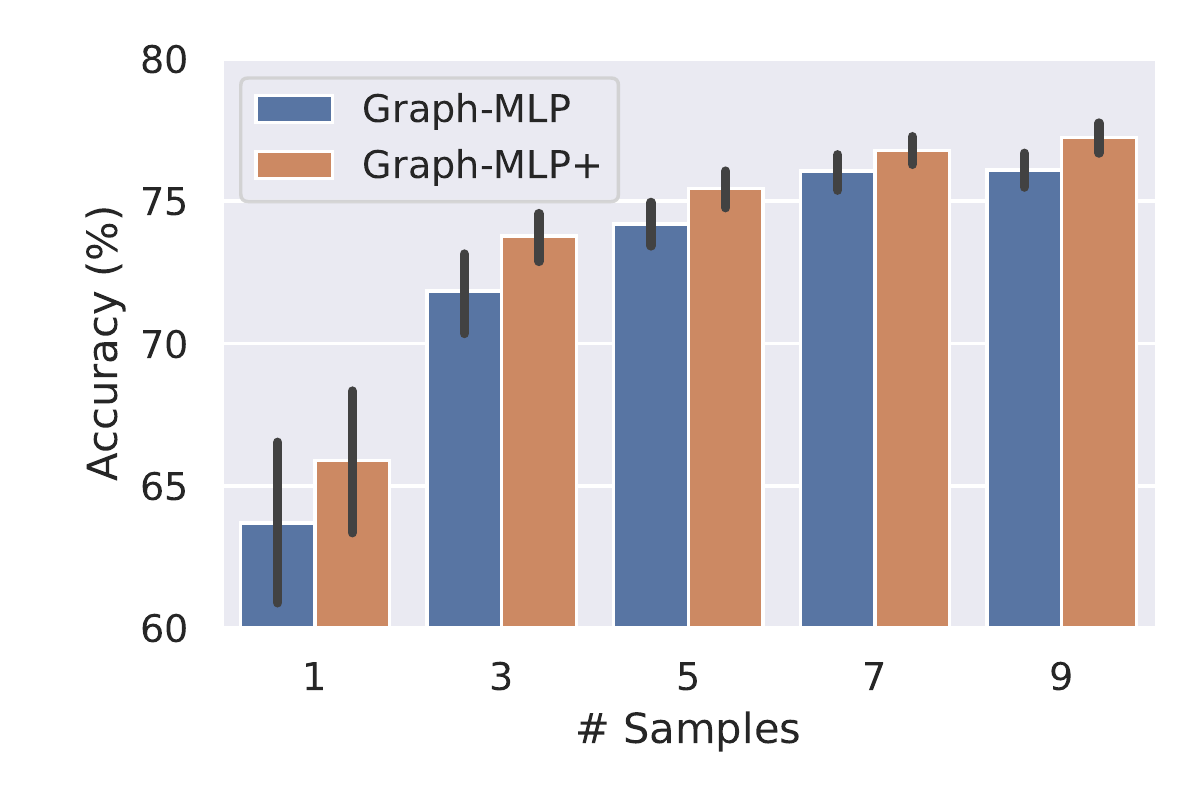}}}%
     \subfloat[Citeseer]{{\includegraphics[width=0.28\linewidth]{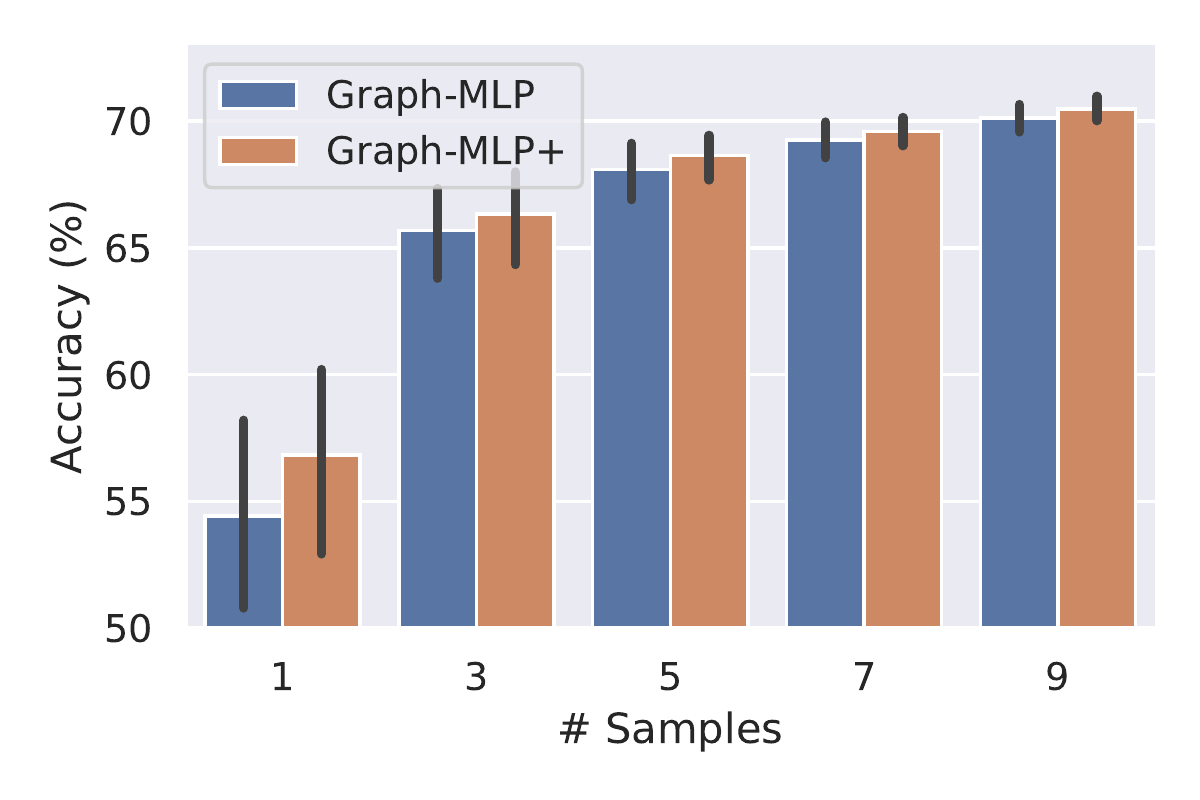} }}%
    \subfloat[Pubmed]{{\includegraphics[width=0.28\linewidth]{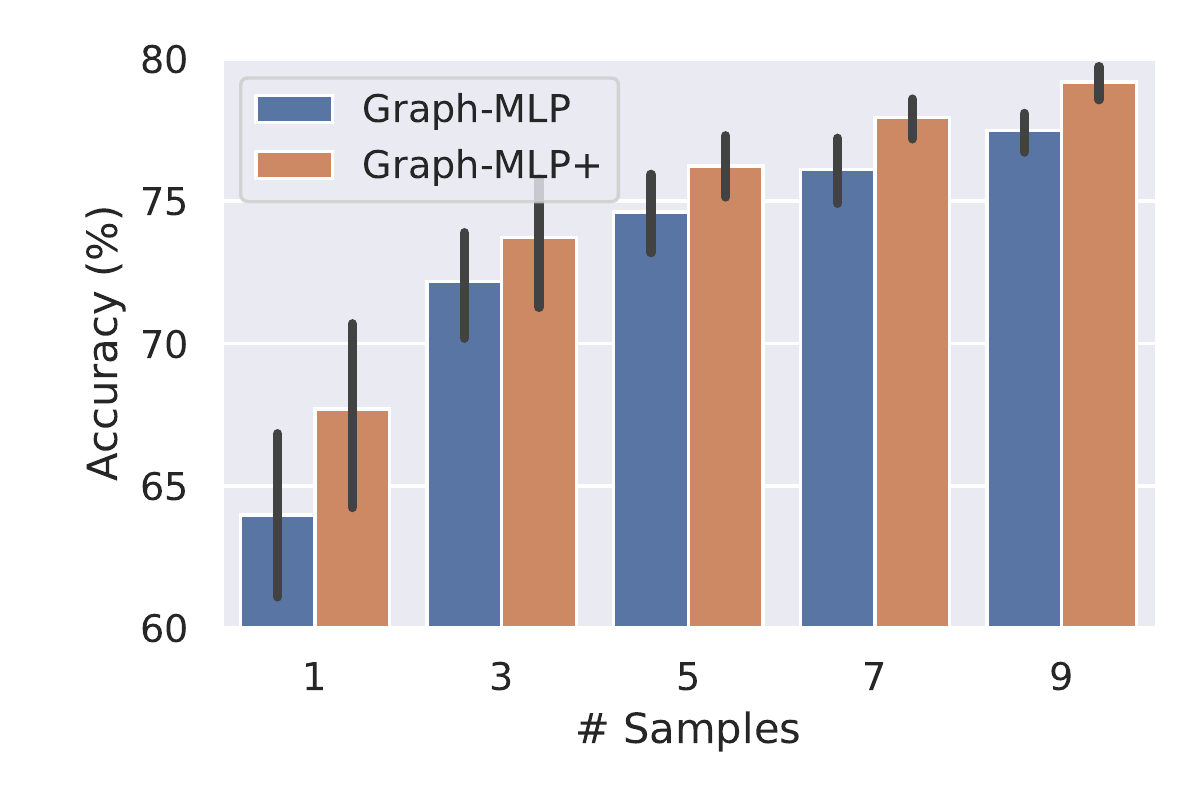} }}%
    \vspace{-3mm}
    \caption{\small{Node classification  results of \gmlp and \gmlpP with limited training samples.}} 
    \label{fig:graphmlp-various-samples}
    \vspace{-1mm}
\end{figure*}
The results of \gra and \graP are summarized in Figure.\ref{fig:grace-mlp-various-samples}. From these figures, it is clear that \graP consistently outperforms \gra on all datasets under various training ratios. These results validate the effectiveness of the proposed enhanced objective. 
\begin{table}[h!tb]
\vspace{-3mm}
\caption{\small{Node classification results of \gca and \gcaP.}}
\vspace{-3mm}
\begin{adjustbox}{width=0.46\textwidth}
\begin{tabular}{ccccc}
\toprule
\textbf{Model}        &  \computer & \photo & \wiki & \cocs\\ 
\midrule

\gca    & 87.49 $\pm$ 0.39 & 92.03 $\pm$ 0.39 & 76.46 $\pm$ 1.30 &92.73 $\pm$ 0.21 \\ 



\gcaP    & \textbf{88.15 $\pm$ 0.40 }  & \textbf{92.52 $\pm$ 0.45} & \textbf{78.64 $\pm$ 0.18} & \textbf{92.82 $\pm$ 0.29}\\ 
 
 
 \bottomrule
\end{tabular}
\end{adjustbox}
\label{tab:gca}
\vspace{-3mm}
\end{table}
\subsubsection{\gca} 
Following~\cite{zhu2021graph}, we evaluate \gca and \gcaP on $4$ datasets including \photo, \computer, \cocs, and \wiki. We follow the same experimental setting as in~\cite{zhu2021graph}. Specifically, we randomly split the nodes into three parts: $80\%$ for testing, $10\%$ for validation, and $10\%$ for training. As in~\cite{zhu2021graph}, we repeat the experiments under $20$ different data splits and report the average performance together with the standard deviation in Table~\ref{tab:gca}. For \gca, we adopt the \gca-DE variant 
since it achieves the best performance overall among the three variants proposed in~\cite{zhu2021graph}. Note that the results of \gca are reproduced using the official code and exact parameter settings provided in~\cite{zhu2021graph}. The performances of \gca and \gcaP are reported in Table~\ref{tab:gca}. As demonstrated in the Table, \gcaP surpasses \gca on most datasets, which further illustrates the effectiveness of the proposed enhanced objective. Also, it suggests that the enhanced objective is general and can be utilized to advance various GCL methods.




\subsection{\gmlp }\label{sec:graph-mlp-results}
In this section, we investigate how the enhanced objective helps improve the performance of \gmlp by comparing its performance with \gmlpP. A brief introduction of \gmlp and \gmlpP can be found in Section~\ref{sec:gmlp-intro}. 

\begin{table}[h!tb]
\caption{\small{Node classification results of \gmlp and \gmlpP.}}
\vspace{-3mm}
\begin{tabular}{ccccc}
\toprule
\textbf{Model}        & \cora & \citeseer & \pubmed \\ 
\midrule
\gmlp   & 79.7$\pm$ 1.15 & 72.99$\pm$0.54 & 79.62$\pm$0.67 \\ 
\gmlpP    & \textbf{80.51$\pm$0.69} & \textbf{74.03$\pm$0.51} & \textbf{81.42$\pm$0.92} \\ 
 \bottomrule
\end{tabular}
\vspace{-3mm}
\label{tab:mlp_public_comp}
\end{table}
Following~\cite{graph_mlp_cite}, we adopt three datasets including \cora, \citeseer, and \pubmed for comparing \gmlpP with \gmlp. As described in Section~\ref{sec:gmlp-intro}, \gmlp runs in a semi-supervised setting. The classification model is trained in an end-to-end way. We adopt the conventional public splits of the datasets~\cite{kipf2016semi} to perform the experiments. Again, the experiments are repeated for $30$ times with different random initialized parameters, and the average performance is reported in Table~\ref{tab:mlp_public_comp}. As shown in Table~\ref{tab:mlp_public_comp}, \gmlpP outperforms \gmlp by a large margin on all three datasets. \gmlpP even outperforms message-passing methods such as GCN by a large margin, especially on \citeseer and \pubmed, which indicates that the proposed objective can effectively incorporate the graph structure information and improve the performance of MLPs. Note that, in the \gmlpP framework, only the MLP model is utilized for inference after training, which is more efficient than message-passing methods in terms of both time and complexity. 

We further compare \gmlpP with \gmlp under the setting where labeled nodes are extremely limited. Specifically, we keep the test and validation set fixed, and only use $s$ samples per class for training, where $s$ is set to $1,3,5,7$, and $9$. When creating these various training sets, the $s$ samples per class are sampled from the training set in the public split setting. The results are presented in Figure~\ref{fig:graphmlp-various-samples}. \gmlpP achieves stronger performance than \gmlp under all settings over all three datasets. Furthermore, \gmlpP performs extremely well when labels are limited. This indicates that the proposed enhanced objective can more effectively utilize the graph structure and feature information, which leads to high-quality node representations even when labels are scarce.  
\vspace{-1mm}
\subsection{Ablation Study}\label{sec:ablation_study}
In this section, we conduct ablation studies to investigate the effectiveness of key components in the proposed enhanced objective. Specifically, we first study how the positive weights $w^+_v(v_j)$ and negative weights $w^-_v(v_j)$ contribute to the overall improvement of the model performance. Then, we conduct another ablation study to investigate how the two types of similarity, i.e,  graph similarity and feature similarity, help model the similarity measure and thus advance the model performance. We only conduct ablation study based on \gra and \gmlp as \gca is a variant of \gra. For \gra, we adopt the $10\%/10\%/80\%$ training, validation and testing split. For \gmlp, we use the conventional public splits (the settings are detailed in Section~\ref{sec:grace} and Section~\ref{sec:graph-mlp-results}, respectively).
\subsubsection{Positive and negative weights in the enhanced objective}
In Section~\ref{sec:enhanced_objective}, two weights $w^+_v(v_j)$ and $w^-_v(v_j)$ are introduced to increase the variety of positive samples and alleviate the effect of false negative samples, respectively. In this section, we aim to investigate how these two kinds of weights contribute to the model performance. For this investigation, we introduce two variants of the proposed objective that only incorporate the positive weights or negative weights. The results for \graP and \gmlpP and their corresponding variants are summarized in Table~\ref{tab:grace_ablation_info} and Table~\ref{tab:mlp_ablation_info}, respectively. Specifically, in Table~\ref{tab:grace_ablation_info}, we denote the variant of \graP with only $w^+_v(v_j)$ as \graPP and the one with only $w^-_v(v_j)$ is denoted as \graPN. Likewise, the two variants for \gmlpP are denoted as \gmlpPP and \gmlpPN in Table~\ref{tab:mlp_ablation_info}. In Table~\ref{tab:grace_ablation_info}, both \graPP and \graPN consistently outperform \gra on all datasets. Similarly, in Table~\ref{tab:mlp_ablation_info}, both \gmlpPP and \gmlpPN consistently outperform \gmlp on all datasets. These results clearly illustrate that both positive weights and negative weights are important for improving the enhanced objectives. Furthermore, these two types of weights contribute to the objectives in a complementary way since \gmlpP outperforms all variants on most datasets. 

\begin{table}[!htbp]
\caption{\small{Node classification results of \graP and its variants.}}
\vspace{-2mm}
\begin{adjustbox}{width=0.43\textwidth}
\begin{tabular}{ccccc}
\toprule
     & \cora &\citeseer & \pubmed  & \dblp\\ 
\midrule
\gra &  82.56$\pm$1.21    & 71.23$\pm$0.86 & 86.12$\pm$0.23  & 84.43$\pm$0.25 \\  
\midrule
\graPP  & 83.59$\pm$0.98   & 71.53$\pm$0.97  & 86.41$\pm$0.29  & 84.76$\pm$0.23 \\ 
\graPN    & 83.20$\pm$1.26 & 72.19$\pm$0.78 & 86.31$\pm$0.22  &   84.89$\pm$0.26 \\ 
\midrule
\graPG  & 83.19$\pm$1.31   & 71.40$\pm$1.03  & 86.34$\pm$0.23  & 84.55$\pm$0.26 \\ 
\graPF    & 82.71$\pm$1.29 & 71.90$\pm$0.87 & 86.11$\pm$0.26  &  84.65$\pm$0.26 \\ 
\midrule
\graP & 83.62$\pm$1.13  & 72.26$\pm$0.82  & 86.45$\pm$0.29 & 84.77$\pm$0.27  \\ 
 \bottomrule
\end{tabular}
\end{adjustbox}
\label{tab:grace_ablation_info}
\vspace{-2mm}
\end{table}

\begin{table}[!htbp]
\caption{\small{Node classification results of \gmlpP and its variants.}}
\vspace{-2mm}
\begin{tabular}{cccc}
\toprule
      & \cora & 
      \citeseer & \pubmed \\ 
\midrule
\gmlp  & 79.7$\pm$1.15 & 72.99$\pm$0.54 & 79.62$\pm$0.67 \\ 
\midrule
\gmlpPP  & 80.33$\pm$0.96  & 73.69$\pm$0.45 & 79.79$\pm$0.89 \\ 
\gmlpPN    & 79.80$\pm$0.86 & 73.41$\pm$0.44 & 79.76$\pm$0.82\\ 
\midrule
\gmlpPG  & 80.32$\pm$0.73  & 73.4$\pm$0.36 & 79.96$\pm$0.94 \\ 
\gmlpPF    & 61.80$\pm$0.61 & 64.04$\pm$0.73 & 75.92$\pm$0.90\\ 
\midrule
\gmlpP & 80.51$\pm$0.69 & 74.03$\pm$0.51 & 81.42$\pm$0.92\\
\bottomrule
\end{tabular}
\vspace{-5mm}
\label{tab:mlp_ablation_info}
\end{table}


\subsubsection{Similarity measure}
 In this part, we aim to investigate how the graph and feature similarity as described in Section~\ref{sec:similarity}. We introduce two variants of the proposed enhanced objectives, one of which only uses graph similarity while the other one only utilizes feature similarity. The results for \graP and \gmlpP and their variants are shown in Table~\ref{tab:grace_ablation_info} and Table~\ref{tab:mlp_ablation_info}, respectively. In the tables, we denote the variant of \graP with only graph similarity information as \graPG, and the one with feature similarity is denoted as \graPF. Similarly, the two variants for \gmlpP are denoted as \gmlpPG and \gmlpPF, respectively. We make the following observations:
\begin{itemize}
    \item In Table~\ref{tab:grace_ablation_info}, both \graPG and \graPF outperforms \gra on most datasets, which demonstrates that both graph and feature similarity contain important information about node similarity and they can be utilized for effectively modeling the anchor-aware distributions. \graP outperforms the two variants and the base model \gra on all datasets, which indicates that the graph similarity and feature similarity are complementary to each other, and properly combining them results in better similarity estimation leading to strong performance.
    \item In Table~\ref{tab:mlp_ablation_info}, \gmlpPG significantly outperforms \gmlp while \gmlpPF does not perform well. This is potentially due to the lack of graph information in MLP models. Different from \graPF which incorporates graph information in the encoder, \gmlpPF only utilizes feature information, which leads to low performance. On the other hand, the strong performance of \gmlpP suggests that the enhanced objective effectively incorporates the graph structure information. However, this does not mean the feature similarity is not important. \gmlpP outperforms \gmlpPG on all three datasets, which suggests that the feature similarity brings additional information than graph similarity, and properly combining them is important.   
\end{itemize}
\vspace{-2mm}

%% file: sections/related_work.tex
\section{Related work}\label{sec:related_work}
 In this section, we review some relevant works.
 
\noindent{\bf Graph Neural Networks.}
 Graph neural networks (GNNs) have achieved great success in generating informative representations from graph-structured data and hence help facilitate many graph-related tasks~\cite{kipf2016semi, velivckovic2017graph,klicpera2018predict, wu2019simplifying}. Graph neural networks often update node representations via a message-passing process, which effectively incorporates the graph information for representation learning~\cite{gilmer2017neural}. More recently, there are attempts to train MLP models that are as capable as GNNs via knowledge distillation~\cite{zhang2022graphless} and neighborhood contrastive learning~\cite{graph_mlp_cite}. 
 
\noindent{\bf Contrastive Learning.} Contrastive learning (CL) aims to learn latent representations by discriminating positive from negative samples. The instance discrimination loss is introduced in \cite{wu2018unsupervised} without data augmentation. In \cite{bachman2019learning}, it is proposed to generate multiple views by data augmentation and learn representations by maximizing mutual information between those views. Momentum Contrast (MoCo) \cite{he2020momentum} maintains a memory bank of negative samples, which significantly increases the number of negative samples used in the contrastive loss calculation. In \cite{chen2020simple}, it is discovered that the composition of data augmentations plays a critical role in CL. BYOL and Barlow Twins \cite{zbontar2021barlow} \cite{grill2020bootstrap} achieves strong CL performance without using negative samples. Recently, a series of tricks such as debiased negative sampling~\cite{chuang2020debiased}, and hard negative mining \cite{kalantidis2020hard, robinson2020contrastive, wu2020conditional}, positive mining \cite{dwibedi2021little} have been proposed and proved effective. Efforts have also been made to extend CL to supervised setting~\cite{khosla2020supervised}. 

\noindent{\bf Graph Contrastive Learning.}
Deep Graph Infomax (DGI) \cite{velickovic2019deep} takes a local-global comparison mode by maximizing the mutual information between patch representations and high-level summaries of graphs. Graphical Mutual Information (GMI) \cite{peng2020graph} maximizes the mutual information between the representation of a node and its neighborhood. MVGRL~\cite{hassani2020contrastive} contrasts multiple structural views of graphs generated by graph diffusion. GRACE~\cite{zhu2020deep} utilizes edge removing and feature masking to generate two different views for node-level contrastive learning. Based upon GRACE, GCA~\cite{zhu2021graph} adopts adaptive augmentations by considering the topological and semantic aspects of graphs. There are also investigations to address the bias induced in the negative sampling process~\cite{zhao2021graph, https://doi.org/10.48550/arxiv.2110.02027}. MERIT \cite{jin2021multi} leverages Siamese GNNs to learn high-quality node representations. Most of these contrastive learning frameworks utilize negative samples in their training. More recently, inspired by BYOL, a framework named BGRL without requiring negative samples was proposed for graph contrastive learning~\cite{thakoor2021bootstrapped}.


 

%% file: sections/conclusion.tex
\section{Conclusion}\label{sec:conclusion}
In this paper, we propose an effective enhanced contrastive objective to approximate the ideal contrastive objective for graph contrastive learning. The proposed objective leverages node similarity to model the anchor-aware distributions for sampling positive and negative samples. Also, the objective is designed to be flexible and general, which could be adopted for any graph contrastive learning framework that utilizes the traditional InfoNCE-based objective. Furthermore, the proposed enhancing philosophy generally applies to other contrasting-based models such as \gmlp which includes an auxiliary contrastive loss. Extensive experiments demonstrate the effectiveness of the enhanced objectives. Some potential future directions include launching other metrics for measuring structure-based node similarity and utilizing parameterized functions to model the node similarity with graph structure and node feature information.
